\documentclass{article}

\usepackage{etoolbox}
\newtoggle{todo}
\newtoggle{arxiv}
\toggletrue{todo}
\toggletrue{arxiv}

\usepackage{microtype}
\usepackage{graphicx}
\usepackage{caption}
\usepackage{subcaption}
\usepackage{booktabs} %

\usepackage{hyperref}
\usepackage[inline]{enumitem}
\usepackage[table]{xcolor}
\usepackage{hhline}
\usepackage{comment}

\iftoggle{arxiv}{
  \usepackage[numbers]{natbib}
  \setlength{\textwidth}{6.5in}
  \setlength{\textheight}{9in}
  \setlength{\oddsidemargin}{0in}
  \setlength{\evensidemargin}{0in}
  \setlength{\topmargin}{-0.5in}
  \newlength{\defbaselineskip}
  \setlength{\defbaselineskip}{\baselineskip}
  \setlength{\marginparwidth}{0.8in}
}{

  \usepackage[accepted]{icml2019}

  \usepackage{natbib}
}

\usepackage{parskip}
\usepackage{xspace}
\usepackage{amsopn,amsmath,amsthm,amssymb}
\usepackage{algorithm}
\usepackage{algorithmic}
\usepackage{breqn}
\usepackage{tabu}
\usepackage{url}
\usepackage{color}
\usepackage{mathtools}
\usepackage{pbox}
\usepackage{scalerel}

\makeatletter
\def\thm@space@setup{%
  \thm@preskip=8pt plus 1pt minus 2pt
  \thm@postskip=\thm@preskip
}
\makeatother
\theoremstyle{definition}
\newtheorem{definition}{Definition}

\theoremstyle{plain}

\newtheorem{proposition}{Proposition}

\newtheorem{conjecture}{Conjecture}

\newcommand{\R}{\mathbb{R}}
\newcommand{\C}{\mathbb{C}}

\newcommand{\diag}[1]{\mathbf{diag}\left( #1 \right)}

\newcommand{\norm}[1]{\left\| #1 \right\|}

\DeclareMathOperator*{\poly}{poly}
\newcommand{\eqdef}{:=}

\newcommand{\F}{\mathbb{F}}

\providecommand{\sign}{\mathop{\rm sign}}
\newcommand\VCdim{{\operatorname{VCdim}}}

\newcommand{\dimmatrix}[2]{\underbrace{\begin{bmatrix}#1\end{bmatrix}}_{\displaystyle #2}}
\newcommand{\bigzero}[1]{\pbox[c][24pt]{24pt}{\normalfont\Large\bfseries 0$_{\scaleto{#1}{4pt}}$}}
\newcommand{\bigeye}[1]{\pbox[c][32pt]{32pt}{\normalfont\Large\bfseries I$_{\scaleto{#1}{4pt}}$}}

\iftoggle{arxiv}{
  \title{Learning Fast Algorithms for Linear Transforms Using Butterfly Factorizations}
  \usepackage{authblk}
  \author[1]{Tri Dao}
  \author[1]{Albert Gu}
  \author[2]{Matthew Eichhorn}
  \author[2]{Atri Rudra}
  \author[1]{Christopher R{\'e}}
  \affil[1]{Department of Computer Science, Stanford University}
  \affil[2]{Department of Computer Science and Engineering, University at Buffalo, SUNY}
  \affil[ ]{\texttt{\{trid,albertgu\}@stanford.edu}, \texttt{\{maeichho,atri\}@buffalo.edu}, \texttt{chrismre@cs.stanford.edu}}
}{
  \icmltitlerunning{Learning Fast Algorithms for Linear Transforms Using Butterfly Factorizations}
}
\date{\today}

\begin{document}

\iftoggle{arxiv}{
  \maketitle
}{
  \twocolumn[
  \icmltitle{Learning Fast Algorithms for Linear Transforms Using Butterfly Factorizations}

  \icmlsetsymbol{equal}{*}

  \begin{icmlauthorlist}
  \icmlauthor{Tri Dao}{Stanford}
  \icmlauthor{Albert Gu}{Stanford}
  \icmlauthor{Matthew Eichhorn}{Buffalo}
  \icmlauthor{Atri Rudra}{Buffalo}
  \icmlauthor{Christopher R{\'e}}{Stanford}
  \end{icmlauthorlist}

  \icmlaffiliation{Stanford}{Department of Computer Science, Stanford University, USA}
  \icmlaffiliation{Buffalo}{Department of Computer Science and Engineering, University at Buffalo, SUNY, USA}

  \icmlcorrespondingauthor{Tri Dao}{trid@cs.stanford.edu}

  \icmlkeywords{Learning algorithms, Fast transforms, Signal processing, Matrix factorization}

  \vskip 0.3in
  ]

  \printAffiliationsAndNotice{}  %

}

\begin{abstract}
  Fast linear transforms are ubiquitous in machine learning, including the
  discrete Fourier transform, discrete cosine transform, and other structured
  transformations such as convolutions.
  All of these transforms can be represented by dense matrix-vector
  multiplication, yet each has a specialized and highly efficient (subquadratic)
  algorithm.
  We ask to what extent hand-crafting these algorithms and implementations is necessary,
  what structural priors they encode, and how much
  knowledge is required to automatically learn a fast algorithm for a
  provided structured transform.
  Motivated by a characterization of matrices with fast matrix-vector multiplication as
  factoring into products of sparse matrices, we introduce a parameterization of
  divide-and-conquer methods that is capable of representing a large class of
  transforms.
  This generic formulation can automatically learn an efficient algorithm for
  many important transforms; for example, it recovers the $O(N \log N)$
  Cooley-Tukey FFT algorithm to machine precision, for dimensions $N$
  up to $1024$.
  Furthermore, our method can be incorporated %
  as a lightweight replacement of generic matrices in machine learning pipelines to learn efficient and compressible transformations.
  On a standard task of compressing a single hidden-layer network,
  our method exceeds the classification accuracy of unconstrained matrices on
  CIFAR-10 by 3.9 points---the first time a structured approach has done so---with 4X faster inference speed and
  40X fewer parameters. %
\end{abstract}

\section{Introduction}
\label{sec:introduction}

Structured linear transformations, such as the discrete Fourier transform (DFT),
discrete cosine transform (DCT), and Hadamard transform, are a workhorse of
machine learning, with applications ranging from data preprocessing, feature
generation, and kernel approximation, to image and language modeling (convolutions).
To date, these transformations rely on carefully designed algorithms, such as the
famous fast Fourier transform (FFT) algorithm, and on specialized
implementations (e.g., FFTW and cuFFT).
Moreover, each specific transform requires hand-crafted implementations for every platform (e.g., Tensorflow and PyTorch lack the fast Hadamard transform), and it can be difficult to know when they are useful.
Ideally, these barriers would be addressed by automatically learning the most
effective transform for a given task and dataset, along with an efficient implementation
of it.
Such a method should be capable of recovering a range of fast transforms with high accuracy and realistic sizes given limited prior knowledge.
It is also preferably composed of differentiable primitives and basic operations common to linear algebra/machine learning libraries, that allow it to run on any platform and be integrated into modern ML frameworks such as PyTorch/Tensorflow.
More fundamentally, this problem ties into the foundational question of understanding the minimal prior knowledge needed to learn high-speed systems, in the spirit of modern trends toward relaxing manually imposed structure (i.e., AutoML).
Recent progress in this vein of learning computational primitives includes addition/multiplication gates~\citep{trask2018neural}, the Strassen $2\times2$ matrix multiplication algorithm~\citep{tschannen2018strassennet}, and PDE solvers~\citep{hsieh2019learning}.

We propose a method that addresses this problem for a class of important transforms that includes the aforementioned examples.
A key challenge lies in defining or parameterizing the space of transforms and corresponding fast algorithms,
which requires using a minimal amount of prior knowledge that captures important and interesting transforms while remaining learnable and efficient.
\citet{egner2001automatic,egner2004symmetry} previously posed this question and a novel combinatorial approach, but their solution only addresses a limited set of transforms (primarily DFT) and only on limited problem sizes.
In particular, these approaches search through an exponentially large discrete space using a symbolic form of the matrix~\cite{egner2001automatic,egner2004symmetry} and recover the solution only up to dimensions $8 \times 8$.
We instead draw two key lessons from the work of \citet{desa2018two},
who characterize matrices with efficient matrix-vector multiplication
algorithms as being factorizable into products of sparse matrices.%
\footnote{This characterization was equivalently known in the language of arithmetic circuits~\citep{burgisser2013algebraic}.}
Thus, the task of learning algorithms can be reduced to finding appropriate
sparse matrix product representations of the transforms.
They further show that divide-and-conquer schemes lead to fast
multiplication algorithms for a surprisingly general set of structured matrices.
Motivated by the broad applicability of this recursive structure, we propose a
particular factorization using sequences of special block diagonal matrices,
called \emph{butterfly matrices}.
Specific instances of butterfly structure have been used before---for example as a random orthogonal
preconditioner~\citep{parker1995random} or
in matrix approximation~\citep{li2015butterfly}---but we use a relaxed representation
that captures a larger class of structures and can learn from data.
These form a class of structured matrices with $O(N)$ parameters and
automatic fast multiplication in $O(N \log N)$ operations.

We empirically validate our method in two ways. %
First, we consider a specification of a transform (e.g., $N$ input-output pairs) and attempt to factorize it.
We successfully recover a fast algorithm up to machine precision for several important transforms such as the DFT,
Hadamard, DCT, and convolution for realistic sizes (dimensions up to $N=1024$), while standard sparse and low-rank baselines cannot (Section~\ref{sec:learning_fast_transforms}).
Beyond recovering famous transforms, we additionally incorporate this method in end-to-end ML pipelines to learn fast and compressible latent transformations (Section~\ref{sec:applications}). %
On the benchmark single hidden layer network, this parameterization exceeds the classification accuracy of a baseline fully connected layer on several datasets---such as by 3.9 points on CIFAR-10 while using 40X fewer parameters---%
which is to our knowledge the first time a structured model has outperformed the unconstrained model for this task on a realistic dataset~\citep{thomas2018learning}.
We also find that the addition of a lightweight butterfly layer improves the accuracy of a modern ResNet architecture by 0.43 points.

Finally, our method is simple with an easily implementable fast algorithm.
We compare the training and inference speed of our implementation
to specialized implementations of discrete transforms (Section~\ref{sec:speed}).
Our generic representation comes within 3-5X of implementations for specific transforms such as the DFT and DCT,
while still being capable of learning a rich class of more general transforms.

\section{Related Work}
\label{sec:related_work}

Fast transforms are crucial and ubiquitous in the machine learning pipelines,
from data preprocessing, feature generation, and dimensionality reduction to
compressing models.
For example, the DFT and DCT form the basis of the mel-frequency cepstral
coefficients (MFCCs), a standard feature representation for speech
recognition~\citep{jurafsky2014speech}.
State-of-the-art kernel approximation methods leverage circulant matrices (i.e.,
convolution)~\citep{yu15} and the DFT and Hadamard
transform~\citep{pmlr-v28-le13, yu2016orthogonal} for fast projection.
Structured matrices, which are matrix representations of fast transforms, play a
crucial role in designing fast neural network layers with few
parameters~\citep{sindhwani2015structured, ding2017circnn}.

Given their importance, there have been significant efforts in finding more and
more general classes of fast transforms.
Traditional classes of structured matrices such as the Toeplitz, Hankel,
Vandermonde, and Cauchy matrices are ubiquitous in engineering and signal
processing~\citep{pan-book}, and more recently have found use in deep learning.
These were generalized under the seminal notion of low displacement rank (LDR)
introduced by~\citet{kailath1979displacement},
\iftoggle{arxiv}{
  and were later unified under a
  single class of displacement structure (the confluent Cauchy-like matrices)
  introduced by~\citet{OS00} to solve the Nevanlinna-Pick interpolation problem.
  Another class of fast transforms that directly generalize the DFT and DCT are
  based on orthogonal polynomials~\citep{chihara}, which find usage in areas from
  differential equations to optics.
  Both orthogonal polynomial transforms~\citep{driscoll}, and all of the
  previously introduced matrices with displacement rank structure, were
  further significantly generalized under a single class by~\citet{desa2018two}.
  Notably, almost all of the structured matrix classes mentioned here exhibit a
  form of recursive structure in their construction and superfast algorithms.
}{
  These, along with more general LDR as well as other families of transforms
  related to the DFT
  and DCT,
  were further significantly generalized under a single class by~\citet{desa2018two}.
  Notably, almost all of the structured matrix classes mentioned here exhibit a
  form of recursive structure.
}

Since the product of sparse matrices immediately has a fast multiplication
algorithm, the problem of sparse matrix factorization has been tackled in many
settings.
Sparse PCA~\citep{zou2006sparse} and dictionary
learning~\citep{mairal2009supervised} factor a matrix into two components, one
of which is sparse.
Sparse matrix factorization with more than two factors has also been considered,
for example in the setting where the true matrix is the product of random sparse
matrices~\citep{neyshabur2013sparse}, or in the context of learning multi-layer
sparse approximations~\citep{lemagoarou2015chasing,lemagoarou2016flexible}.
Our approach differs from these in that we focus on the recursive structure of
the transforms\iftoggle{arxiv}{---not just the sparsity of their factors---}{, }leading to sparse \emph{and} structured transforms, and avoiding the discreteness problem inherent to learning sparsity.

Since most distinct transforms typically require significant work both to design fast
algorithms and to efficiently implement them on different platforms, there have
been attempts to automatically learn these fast algorithms.
The field of algebraic signal processing~\citep{puschel2008algebraic} uses
methods from representation theory of groups and algebras to automatically
generate fast algorithms from the symbolic form of the transform matrix.
However, these methods require search over a combinatorially-large discrete space, limiting their approaches to small matrices of size up to
$8 \times 8$~\citep{egner2004symmetry,voronenko2009algebraic}.
Attempts to learn general algorithms such as matching~\citep{mena2018learning},
sorting~\citep{grover2019stochastic}, and traveling
salesman~\citep{bello2016neural} using differentiable architectures face a
similar challenge of having to effectively explore a large discrete space.
Thus, they only work for problems of size at most 100.
By contrast, our approach
simplifies the discreteness of the problem into learning a simpler set of permutations,
allowing us to recover fast algorithms for realistic dimensions.

Independently, there has been growing interest in compressed deep learning models, motivated by the goal of adapting them to resource-constrained environments.
A common approach for learning compressed models involves replacing the
unconstrained weight matrices with a class of structured matrices and learning directly on the parametrization of that class.
The most effective methods use matrix classes that are explicitly related to Fourier transforms~\cite{sindhwani2015structured}, or employ highly specialized and complicated recursive algorithms~\cite{thomas2018learning}.
As our method also implicitly defines a highly compressible subclass of matrices
with linear parameter count and efficient multiplication, it can be used as a drop-in replacement for matrices in such end-to-end ML models.

\section{Recovering Fast Transforms}
\label{sec:method}

We now set up and describe our approach.
We first reiterate the connection between fast algorithms and sparse matrix factorization,
and briefly outline a quintessential divide-and-conquer algorithm (the FFT) as motivation.

We then elaborate the details of our method for learning particular recursive algorithms,
including a core permutation-learning step that enables it to capture a wider range of structures.
We also discuss the expressive power of these matrices, including which transforms they capture perfectly,
and define a hierarchy of matrix classes built on butterflies that can theoretically capture richer recursive structures.

\subsection{Preliminaries}
\label{sec:preliminaries}

\paragraph{Sparse factorizations}

One method of constructing matrices with obvious fast matrix-vector multiplication is as a product of sparse matrices,
so that multiplication by an arbitrary vector will have cost proportional to the
total number of nonzeros of the matrices in the product.

Surprisingly, the converse is also true.
The notion of \emph{sparse product width} (SPW)~\citep{desa2018two}, which roughly corresponds to the total sparsity of a factorization of a matrix,
turns out to be equivalent to the length of the shortest linear straight-line program describing a matrix (up to a constant).
Hence, it is an optimal descriptor of the algorithmic complexity of matrix-vector multiplication on these types of models~\cite{burgisser2013algebraic}.

\label{subsec:dft}

Given the general correspondence between sparse factorization and fast algorithms, we consider specific types of discrete
transforms and their recursive factorizations.
This is a prototype for our parameterization of fast recursive algorithms in
Section~\ref{sec:fast_transforms}.

\begin{figure}[ht]
  \centering
  \includegraphics[width=0.5\textwidth]{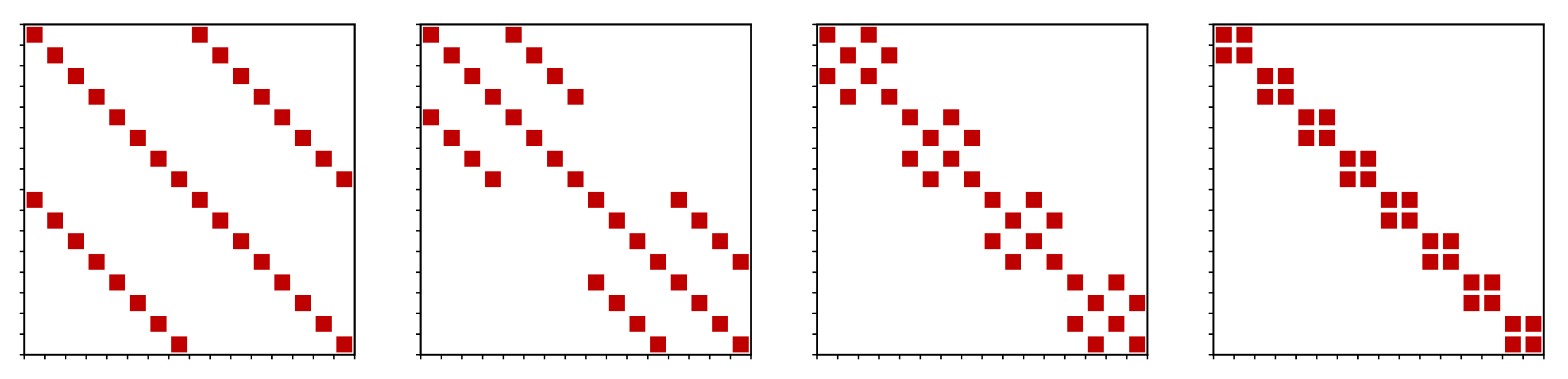}
  \caption{\label{fig:butterfly_matrices} Butterfly matrix for $N = 16$.
    From left to right: single copy of $B_{16}$, blocks of $B_8$, blocks of $B_4$, blocks of $B_2$.}
\end{figure}

\paragraph{Case study: DFT}
The Discrete Fourier Transform (DFT) transforms a complex input vector $x = [x_0, \dots, x_{N-1}]$ into a complex
output vector $X = [X_0, \dots, X_{N-1}]$ by expressing the input in the basis
of the complex exponentials:
\begin{equation*}
  X_{k} = \sum_{n=0}^{N-1} x_n e^{-\frac{2\pi i}{N}kn}, \quad k = 0, \dots, N-1, N = 2^m.
\end{equation*}
Let $\omega_N \eqdef e^{2\pi i/N}$ denote a primitive $N$-th root of unity.
The DFT can be expressed as matrix multiplication by the \emph{DFT matrix} $F_N \in \mathbb{C}^{N \times N}$, where $(F_N)_{kn} = \omega_N^{-kn}$. %
The DFT of size $N$ can be reduced to two DFTs of size $N/2$ on the even
indices and the odd indices:
\begin{equation*}
  F_N x = \begin{bmatrix} F_{N/2} x_{\mathrm{even}} + \Omega_{N/2} F_{N/2} x_{\mathrm{odd}} \\ F_{N/2} x_{\mathrm{even}} - \Omega_{N/2} F_{N/2} x_{\mathrm{odd}} \end{bmatrix},
\end{equation*}
where $x_{\mathrm{even}} = \left[ x_0, x_2, \dots, x_{N-2} \right]$,
$x_{\mathrm{odd}} = \left[ x_1, x_3, \dots, x_{N-1} \right]$, and $\Omega_{N/2}$ is
the diagonal matrix with entries $1, \omega_N^{-1}, \dots, \omega_N^{-(N/2 - 1)}$.
This recursive structure yields the efficient recursive Cooley-Tukey Fast Fourier Transform (FFT) algorithm.
This computation can be written as a matrix factorization
\begin{equation*}
  F_N = \begin{bmatrix} I_{N/2} & \Omega_{N/2} \\ I_{N/2} & -\Omega_{N/2} \end{bmatrix}
  \begin{bmatrix} F_{N/2} & 0 \\ 0 & F_{N/2} \end{bmatrix}
  \begin{bmatrix} \text{ Sort the even } \\ \text{ and odd indices } \end{bmatrix},
\end{equation*}
where $I_{N/2}$ is the identity matrix, and the last factor is the
permutation matrix $P_N$ that separates the even and odd indices (e.g., mapping
$[0, 1, 2, 3]$ to $[0, 2, 1, 3]$) (see Figure~\ref{fig:permutation}).
Unrolling the recursion, we obtain:
\iftoggle{arxiv}{
\begin{equation}
  \label{eq:fft-factorization}
  \begin{aligned}
    F_N =&\ B_N \begin{bmatrix} F_{N/2} & 0 \\ 0 & F_{N/2} \end{bmatrix} P_N \\
    =&\ B_N \begin{bmatrix} B_{N/2} & 0 \\ 0 & B_{N/2} \end{bmatrix}
      \begin{bmatrix} F_{N/4} & 0 & 0 & 0 \\ 0 & F_{N/4} & 0 & 0 \\ 0 & 0 & F_{N/4} & 0 \\ 0 & 0 & 0 & F_{N/4} \\ \end{bmatrix}
     \begin{bmatrix} P_{N/2} & 0 \\ 0 & P_{N/2} \end{bmatrix} P_N \\
    =&\ \cdots \\
    =&\ \left( B_N \dots \begin{bmatrix} B_{2} & \dots & 0 \\ \vdots & \ddots & \vdots \\ 0 & \dots & B_2 \end{bmatrix} \right)
      \left( \begin{bmatrix} P_{2} & \dots & 0 \\ \vdots & \ddots & \vdots \\ 0 & \dots & P_2 \end{bmatrix} \dots P_N \right).
    \end{aligned}
  \end{equation}
  } {
    \begin{align}
    & F_N =\ B_N \begin{bmatrix} F_{N/2} & 0 \\ 0 & F_{N/2} \end{bmatrix} P_N \nonumber \\
    =&\ B_N \begin{bmatrix} B_{N/2} & 0 \\ 0 & B_{N/2} \end{bmatrix}
      \begin{bmatrix} F_{N/4} & 0 & 0 & 0 \\ 0 & F_{N/4} & 0 & 0 \\ 0 & 0 & F_{N/4} & 0 \\ 0 & 0 & 0 & F_{N/4} \\ \end{bmatrix} \nonumber \\
    & \hspace{2em} \begin{bmatrix} P_{N/2} & 0 \\ 0 & P_{N/2} \end{bmatrix} P_N \label{eq:fft-factorization} \\
    =&\ \cdots   \nonumber \\
    =&\ \left( B_N \dots \left[  \begin{smallmatrix} B_{2} & \dots & 0 \\ \vdots & \ddots & \vdots \\ 0 & \dots & B_2 \end{smallmatrix} \right] \right)
    \left( \left[ \begin{smallmatrix} P_{2} & \dots & 0 \\ \vdots & \ddots & \vdots \\ 0 & \dots & P_2 \end{smallmatrix} \right] \dots P_N \right). \nonumber
    \end{align}
  }
The product of all the $B_{N/2^k}$ matrices on the left is called a \emph{butterfly
matrix}, and each factor $B_{N/2^k}$ is a $2\times2$ block matrix of diagonal matrices called a \emph{butterfly factor}.
Figure~\ref{fig:butterfly_matrices} illustrates the sparsity pattern of the structured butterfly factors.
One can also combine the product of permutation matrices on the right to obtain
a single permutation
called the \emph{bit-reversal permutation}, which sorts the indices by the reverse
of their binary representation (e.g.\ $[0, \dots, 7] \to [0, 4, 2, 6, 1, 5, 3, 7]$).

Other transforms have similar recursive structure but differ in the entries of
$B_{N/2^k}$, and in the permutation.
For example, the DCT involves separating the even and the
odd indices, and then reversing the second half (e.g.,
$[0, 1, 2, 3] \to [0, 2, 1, 3] \to [0, 2, 3, 1]$).

Appendix~\ref{sec:examples} provides some examples of how important transforms, such as the DFT, DCT,
Hadamard, and convolutions, can factor as similar products of sparse matrices.

\subsection{Recovering Fast Transform Algorithms}%
\label{sec:fast_transforms}

Many previous works attempt to compress generic matrices by sparsifying them.
We note that allowing
for products of matrices with a total sparsity budget is strictly more
expressive than a single matrix with that sparsity, while retaining the same compression and computation complexity.
Therefore one can hope to recover all fast algorithms by learning over the set
of matrix products with a total sparsity budget.
However, this is infeasible to learn due to the discreteness of
the sparsity constraint (Section~\ref{sec:introduction},\ref{sec:related_work}).
We instead use a class of matrices built as products of
specific factors that captures the recursive nature of many fast
algorithms.

\paragraph{A butterfly parametrization}

Let $x = [x_0, \dots, x_{N-1}]$ be an input vector.%
\footnote{For simplicity, we assume that $N$ is a power of 2.
  Otherwise, the input can be padded with zeros.}
Let $\mathcal{T}_N$ be a linear transform of size $N$ with matrix representation 
$T_N \in \mathbb{F}^{N \times N}$,
where $\mathbb{F} \in \{\R,\C\}$.
A general recursive structure is to separate the input vector into two halves by
some permutation, apply the transform on each half, and combine the result in a
linear manner by scaling by an diagonal matrix and adding the results.
Written as a matrix factorization:
\begin{equation*}
  T_N = \begin{bmatrix} D_1 & D_2 \\ D_3 & D_4 \end{bmatrix} \begin{bmatrix} T_{N/2} & 0_{N/2\times N/2} \\ 0_{N/2 \times N/2} & T_{N/2} \end{bmatrix} P_{N},
\end{equation*}
where $P_N$ is some permutation matrix and $D_1, \dots, D_4 \in \mathbb{F}^{N/2}$
are diagonal matrices.
Inspired by the factors of the FFT, we call the matrix
$\begin{bmatrix} D_1 & D_2 \\ D_3 & D_4 \end{bmatrix}$ a butterfly factor,
denoted by $B_{N}$.
Unrolling the recursion as in equation~\eqref{eq:fft-factorization} gives the factorization
$T_N = B^{(N)} P^{(N)}$, where $B^{(N)}$ is a butterfly matrix and $P^{(N)}$ is a permutation
that can be written as the product of $\log_2(N)$ simpler block permutations.
We also consider composing this module, hence learn either
\begin{equation}
  \label{eq:BP}
  T_N = B^{(N)}P^{(N)} \qquad T_N = B_2^{(N)}P_2^{(N)}B_1^{(N)}P_1^{(N)},
\end{equation}
which we term the BP and the BPBP parametrization respectively.
One dimensional convolutions (i.e.\ circulant matrices) are notably captured by BPBP,
since they can be computed via an FFT, a component-wise product, then an inverse FFT (see Appendix~\ref{sec:examples}).

\iftoggle{arxiv}{
In the case of the FFT, as in Section~\ref{sec:preliminaries}, the entries of the
butterfly factors are called twiddle factors, and the combined permutation
$P^{(N)}$ is called the bit-reversal permutation.
}{}

\paragraph{Learning a recursive permutation}

\begin{figure}[h!]
  \centering
  \includegraphics[width=0.5\textwidth]{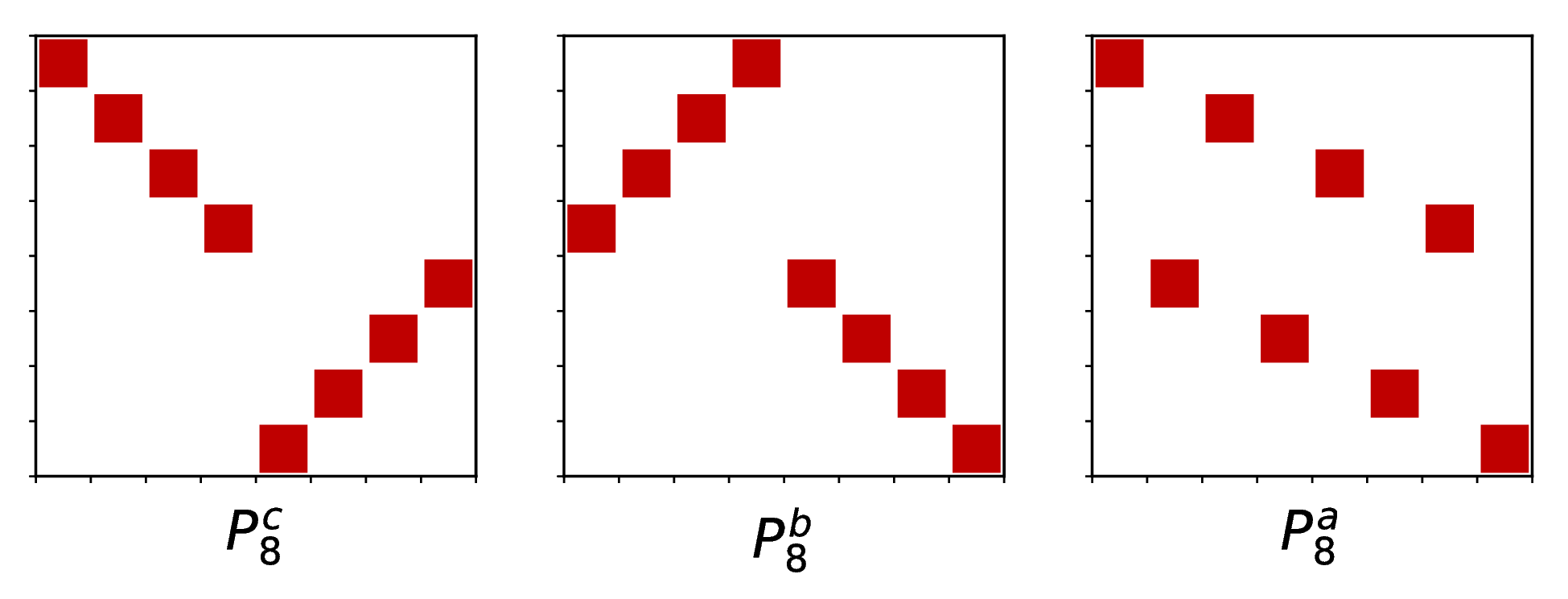}
  \caption{Three binary choices for constructing the permutation used at every step of the recursive process. One of 8 possible permutations can be constructed by multiplying a subset of these matrices in the presented order.} 
  \label{fig:permutation}
\end{figure}

The butterfly blocks in the BP or BPBP parametrization have a fixed sparsity pattern
and their parameters can be directly optimized.
However, the transforms we are interested in capturing frequently require different permutations as part of the ``divide'' step,
which form a set of discrete objects that we must consider.
We will restrict to learning over permutations that have a simple structure
often encountered in these algorithms:
we assume that the distribution factors into $\log_2 N$ steps
following the $\log_2 N$ recursive layers.
At each step in the recursion, the permutation $P_{N/2^k}$ is allowed to either keep
the first half and second half intact or separate the
even and the odd indices (e.g., $[0, 1, 2, 3] \to [0, 2, 1, 3]$). Then, it
can choose to reverse the first half (e.g., $[0, 1] \to [1, 0]$) and can choose
to reverse the second half
(e.g., $[2, 3] \to [3, 2]$).
Thus at each step, there are 3 binary choices and hence 8 possible permutations.
These are illustrated in Figure~\ref{fig:permutation}, where $P_N^a$ denotes the permutation matrix
on $N$ elements that separates the even and odd elements, $P_N^b$ denotes the permutation matrix that reverses the first half,
and $P_N^c$ denotes the permutation matrix that reverses the second half.

Instead of searching over $8^{\log_2 N}$ discrete permutations, we parameterize
the permutation $P^{(N)}$ as a categorical distribution of these $8^{\log_2 N}$
permutations.
The permutation $P_{N/2^k}$ at step $k$ is thus chosen as a convex combination of the $8$ possible choices:
\begin{equation*}
  P_{N/2^k} = p_{cba} P_{N/2^k}^cP_{N/2^k}^bP_{N/2^k}^a + p_{cb}P_{N/2^k}^cP_{N/2^k}^b + \dots.
\end{equation*}
This can be learned by representing this probability distribution $\{p_{cba}, p_{cb},\dots\}$ for example via logits and the softmax.

We make the further simplification that the probabilities $p_{cba}$ factor into the three components; conceptually, that the choices of choosing $P_{N/2^k}^c, P_{N/2^k}^b, P_{N/2^k}^a$ to be part of the product are independent of each other.
This results in the representation
\iftoggle{arxiv}{
\begin{equation}
  \label{eq:permutation}
  P_{N/2^k} = \prod_{s=c,b,a} (p_s P_{N/2^k}^s + (1-p_s)I). %
\end{equation}
Thus we learn the permutation $P_{N/2^k}$ via equation~\eqref{eq:permutation} by optimizing over $3$ logits $\ell_a, \ell_b, \ell_c$ and setting $p_s = \sigma(\ell_s)$, where $\sigma$ is the sigmoid function.
}{
    $P_{N/2^k} = \prod_{s=c,b,a} (p_s P_{N/2^k}^s + (1-p_s)I)$.
  Thus we learn $P_{N/2^k}$ by optimizing over $3$ logits $\ell_a, \ell_b, \ell_c$ and setting $p_s = \sigma(\ell_s)$, where $\sigma$ is the sigmoid function.
}

To encourage the distribution over permutations to be peaked, one can add
entropy regularization~\citep{grandvalet2005semi} or semantic loss~\citep{xu2018a}.
However,
we found that these tricks are not necessary.
For example, the learned transforms in
Section~\ref{sec:learning_fast_transforms} typically put weight at least $0.99$
on a permutation.

\paragraph{Initialization}

As the BP or BPBP construction is a product of many matrices, proper initialization is crucial to avoid exponential blowup in the size of the entries or condition numbers (i.e., the exploding/vanishing gradient problem~\citep{pascanu2013difficulty}).
We aim to initialize each butterfly factor to be close to unitary or orthogonal,
so that the magnitude of the inputs and outputs to the transform are preserved.
This is easy since each of the factors $B_N, \dots, B_2$ has exactly two
nonzeros in each row and column;
for example in the real case, initializing each entry of $B_k$ as
$\mathcal{N}(0,1/2)$ guarantees 
$ \mathbb{E} B_k^* B_k = I_N$.

\paragraph{Comparison to related methods}
Some previous works have examined similar butterfly matrices in numerical algebra or machine learning~\citep{parker1995random,jing2017tunable,munkhoeva2018quadrature}, mainly motivated by trying to parametrize cheap orthogonal matrices.
Our parametrization, motivated by the goal of learning recursive transforms, differs in several ways from all previous works:
\begin{enumerate*}[label=\arabic*.]
  \item We explicitly model and learn a permutation matrix $P$.
  \item Our relaxation does not enforce the matrix to be orthogonal.
  \item Our butterfly factors are ordered so that closer elements interact first
  (Figure~\ref{fig:butterfly_matrices}), whereas some works (e.g.\ \citep{munkhoeva2018quadrature}) reverse the order.
  \item Every work has a different weight-tying scheme; ours ties the blocks in each butterfly factor, leading to fewer parameters and a tighter recursive interpretation than for example~\citep{jing2017tunable}.
\end{enumerate*}

\iftoggle{arxiv}{
  Our main baseline for deep learning experiments is~\citet{thomas2018learning}, who define a special matrix class with a complicated recursive algorithm.
  While our BP method and theirs share some overlap (e.g., they both capture circulant matrices), they have a distinct parametrization, and the exact relation between the BP hierarchy and their LDR-SD or LDR-TD classes is unknown.
  From a practical standpoint, BP is significantly faster
  and simpler to implement than their methods.
}

\subsection{Expressivity and the butterfly hierarchy}

The butterfly matrix $B$ has a total of $4N$ learnable parameters (the
butterfly factors $B_N$, $B_{N/2}$, ..., $B_2$ have $2N$, $N$, ..., $4$ entries respectively).
The overall permutation $P$ has $3\log_2 N$ learnable parameters;
we can also tie the logits of the $\log_2 N$ probabilistic permutations---%
reflecting the fact that for some algorithms the reduction from size $N$ to
$N/2$ is self-similar to the reduction from size $N/2^k$ to $N/2^{k+1}$---%
reducing this to just $3$ parameters.

We can define a natural hierarchy of matrix classes built on the BP primitive. This hierarchy covers a spectrum ranging from extremely structured matrices with a linear number of parameters, to the entire space of square matrices.

\begin{definition}
  For any dimension $N$, let (BP)$^{k}_r$ ($k, r \in \mathbb{N}$) denote the classes of matrices that can be expressed as
  \label{def:bp-hierarchy}
  \iftoggle{arxiv}{
    \[
      S \left( \prod_{i=1}^k B_{i}P_i \right) S^T,
    \]
  }{
    \[
      S \left( \prod_{i=1}^k B_{i}P_i \right) S^T,
    \]
  }
  where each $B_iP_i \in \F^{rN \times rN}$ is a BP module as in equation~\eqref{eq:BP},
  and $S \in \F^{N \times rN} = \begin{bmatrix} I_N & 0 & \hdots & 0 \end{bmatrix}$ (that is, $S$ and $S^T$ select the upper left $N \times N$ entries of the BP product matrix).
  The subscript $r$ is understood to be $1$ if omitted.
\end{definition}
Note that the BP and BPBP classes are equivalent to (BP)$^1$ and (BP)$^2$ respectively.
We remark that $B$ and $P$ are both capable of being the identity, and thus (BP)$^k \subseteq$ (BP)$^{k+1}$.

The BP hierarchy is expressive enough to theoretically represent many important transforms with low depth, as well as all matrices with linear depth:

\begin{proposition}
  \label{prop:expressivity}
  (BP)$^1$ captures the fast Fourier transform, the fast
  Hadamard transform, and their inverses exactly.
  (BP)$^2$ captures the DCT, DST, and convolution exactly. All $N \times N$ matrices are contained in (BP)$^{4N+10}_2$.
\end{proposition}

Proposition~\ref{prop:expressivity} is shown in Appendix~\ref{sec:proofs}.
We suggest some additional conjectures about the expressiveness of the BP hierarchy in Appendix~\ref{sec:bp-conj}.

Even though the BP parameterization is expressive, it still retains the
learnability characteristic of compressed parameterizations.
In fact, neural networks comprising layers of BP and BPBP matrices still have VC
dimension that is almost linear in the number of parameters
(Appendix~\ref{sec:proofs}), similar to networks with fully-connected
layers~\citep{bartlett1999almost,bartlett2017nearly} and
LDR~\cite{thomas2018learning}, which implies a corresponding sample complexity
bound.

\section{Empirical Evaluation}

We evaluate the proposed approach to verify that our butterfly parameterization
can both recover fast transforms and be integrated as an effective component in ML pipelines\footnote{Code to reproduce experiments and plots is available at \small{\url{https://github.com/HazyResearch/butterfly}}}.
In Section~\ref{sec:learning_fast_transforms}, we confirm that it automatically learns the fast algorithms for many discrete transforms commonly used in signal processing and machine learning.
Section~\ref{sec:applications} further shows that it can be a useful component to increase the performance of deep learning models while ensuring fast multiplication and few parameters by design.

\subsection{Discrete Transforms}
\label{sec:learning_fast_transforms}

Below we list several important classes of structured matrices.
Some of them are directly captured by our parametrization and we expect that they can be recovered close to perfectly,
thus providing a $O(N \log N)$ algorithm that closely approximates the naive $O(N^2)$ matrix multiplication.
Others are not perfectly captured by the BPBP class but still have recursive structure; for these, we expect that
our method reconstructs them better than standard matrix compression methods (sparse, low-rank, and combinations) can.

\paragraph{Transforms}
We describe the matrices we evaluate on and their applications;
a standard reference is~\citet{proakis2001digital}.
Their explicit formulas are in Appendix~\ref{sec:examples}, Table~\ref{tab:formulas}.
\begin{enumerate}[topsep=1pt, itemsep=-1ex, partopsep=1ex, parsep=1ex]
  \item Discrete Fourier transform (DFT): arguably the most important
  computational tool in signal processing, the FFT is one of the top 10
  algorithms of the 20th century~\citep{dongarra2000guest}.
  \item Discrete cosine transform (DCT): it expresses the input vector in the
  basis of cosine functions.
  It finds use in lossy compression of audio (MP3) and images (JPEG), in speech
  processing, and in numerical methods of solving partial differential equations
  (PDEs).
  \item Discrete sine transform (DST): similar to the DCT, it expresses the
  input vector as a linear combination of sine functions.
  It is widely employed in spectral methods to solve PDEs.
  \item Convolution: widely used in statistics, image processing, computer
  vision, and natural language processing.
  \item Hadamard transform: commonly used in quantum information processing algorithms,
  and in ML as a fast random projection or kernel approximation method.
  \item Discrete Hartley transform: similar to the DFT, but it transforms real inputs to real outputs.
  It was designed
  as a more efficient option than the DFT for real data.
\end{enumerate}

\paragraph{Methods}
We assume that the transform $\mathcal{T}$ is fully-specified, e.g., from $N$ linearly independent input-output pairs from which the matrix representation $T_N \in \F^{N \times N}$ can be computed.

To recover a fast algorithm of the transform, we wish to approximate $T_N$ with the product of one or more blocks of butterfly and permutation
products, by minimizing the Frobenius norm of the difference:
\begin{equation}
  \mbox{minimize} \quad \frac{1}{N^2} \norm{T_N - B^{(N)} P^{(N)}}^2_F.
  \label{eq:frob_objective}
\end{equation}
By design, this factorization yields a fast $O(N \log N)$ algorithm for the
transform.

We also compare to standard baselines for matrix factorization, maintaining the same total sparsity budget (i.e.\ computation cost of a multiplication) for each:
\begin{enumerate}[topsep=0pt, itemsep=-1ex, partopsep=1ex, parsep=1ex]
  \item Sparse:
  this is the same as choosing the largest $s$ entries where $s$ is the sparsity
  budget.
  \item Low-rank: the sparsity budget is used in the parameters of the low-rank factors, which can be found with a truncated SVD.
  \item Sparse + low-rank: $\norm{T_N - S - L}^2$ is minimized, where $S$ is
  sparse and $L$ is low-rank, by solving a convex problem.\footnote{Although there is an extra addition, this can also be written as a sparse product of 3 matrices by adding auxiliary identity blocks.}
  This is commonly known as robust PCA~\citep{candes2011robust}.
\end{enumerate}

\paragraph{Experimental procedure}
We use the Adam optimizer~\citep{Kingma2014} to minimize the Frobenius norm of the
error, and use Hyperband~\citep{li2017hyperband} to automatically tune the
hyperparameters (learning rates, random seed for initialization).
The runs are stopped early if the average per entry difference (aka RMSE)
$\frac{1}{N} \norm{T_N - B^{(N)} P^{(N)} }_F$ is low enough:
we consider RMSE below 1e-4 (corresponding to the objective
in~\eqref{eq:frob_objective} below 1e-8, while we use 32-bit floats with
machine epsilon around 6e-8) to mean that we successfully recover the fast
algorithms for the transforms to machine precision.
For consistency, we consider the unitary or orthogonal scaling of these
transforms such that they have norm on the order of 1.0.
For the DCT and DST, we add another simple permutation for extra learnability.
All transforms considered learn over BP except for convolution which uses BPBP.
All methods are optimized over complex entries.

Since the forward mapping of our butterfly parameterization is differentiable
with respect to the entries of the butterfly matrices and the logits of the permutations, gradients are easily
obtained with the help of an auto-differentiation framework.
We provide our code in PyTorch.

\begin{figure*}[ht]
  \centering
  \includegraphics[width=0.9\linewidth]{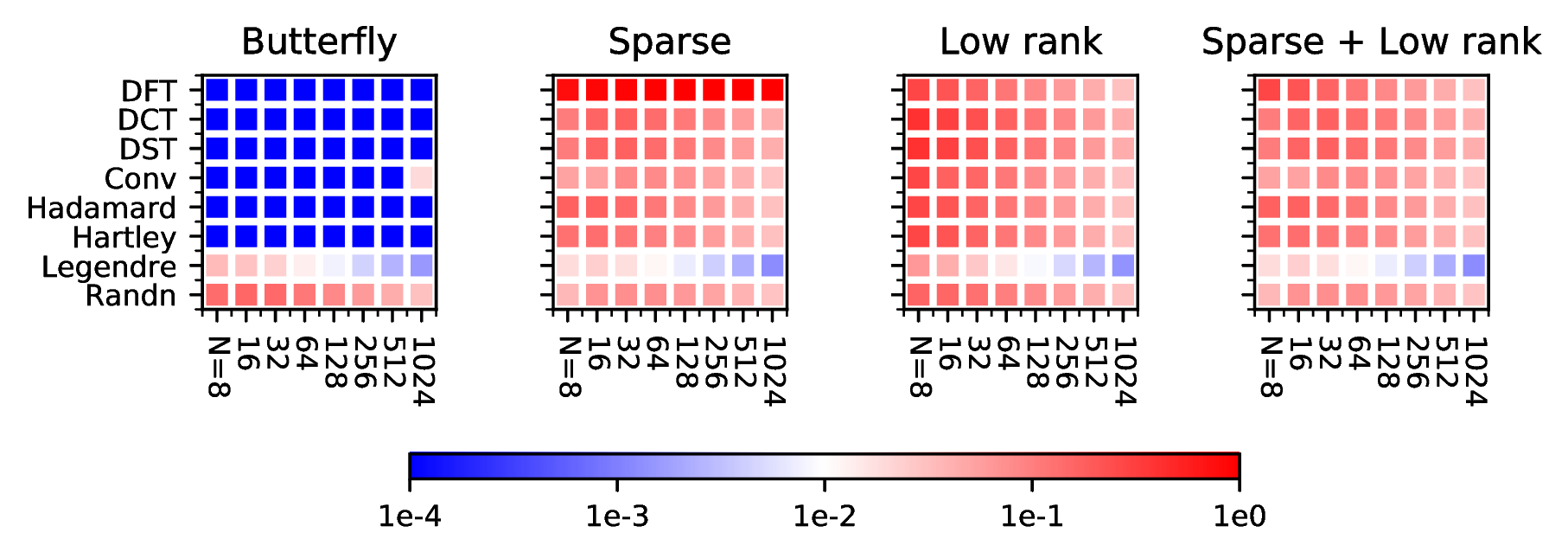}
  \caption{RMSE of learning fast algorithms for common transforms, with early
    stopping when RMSE is below 1e-4. (Blue is better and red is worse.) Our butterfly parameterization can recover common
    transforms up to $N = 1024$ and convolutions up to $N = 512$. Explicit formulas for each transform are listed in Appendix~\ref{sec:examples}, Table~\ref{tab:formulas}.}
  \label{fig:learning_fast_transforms}
\end{figure*}

\paragraph{Quality}
Figure~\ref{fig:learning_fast_transforms} visualizes the lowest error found by Hyperband for various matrix dimensions and several methods.
Full numerical results are provided in Appendix~\ref{sec:extraexps}.
As shown, we successfully
recover the fast algorithms for these transforms up to $N=512$ for convolution
and $N=1024$ for other transforms.
For example, the matrix factorization procedure recovers the bit-reversal
permutation applied at the beginning of the Cooley-Tukey fast Fourier transform.
It also discovers many other unconventional permutations that also lead to exact factorization
of the FFT.

We note that there are other transforms not captured by our parameterization.
Orthogonal polynomial transforms, such as the discrete Legendre transform (DLT),
are known only to have fast $O(N \log^2 N)$ algorithms.
They follow a slightly more general divide-and-conquer decomposition that we
elaborate on in Appendix~\ref{subsec:ops}.
As expected, we find that the butterfly parameterization does not perfectly capture the DLT,
but does recover it slightly better than the baselines. %

Figure~\ref{fig:learning_fast_transforms} also includes a baseline row factoring a matrix of
appropriately scaled i.i.d.\ Gaussian entries,
to indicate typical errors for factoring an unstructured matrix.

\subsection{Neural Network Compression}
\label{sec:applications}

Many structured matrix approaches have been proposed to replace fully-connected (FC)
layers of neural networks, to speed up training and inference, and to reduce the
memory consumption.
These structured matrices are cleverly designed by combining commonly used fast
transforms.
For example, Fastfood~\citep{pmlr-v28-le13} and Deep Fried Convnets~\citep{yang2015deep} compose the fast Hadamard transform and
fast Fourier transforms, %
and \citet{sindhwani2015structured} use Toeplitz-like matrices that can be written
as a sequence of 2 or 4 FFTs.
However, the design choice for these light-weight replacement layers is restricted by the set of
known and implementable transforms.

On the first benchmark task of compressing a single hidden layer model, the real version of BPBP has better classification accuracy than a fully-connected layer on all datasets tested, and uses more than 56X fewer parameters (Table~\ref{table:images}); the complex version performs even better with a slight parameter increase.
The previous best methods fail to achieve this on the more challenging CIFAR-10 dataset at the same parameter budget~\citep{thomas2018learning}.
We further demonstrate that this layer is effective as a lightweight addition to a larger-scale ResNet architecture.

\paragraph{Fully-connected}

Previous work showed that structured matrix approaches based on the low displacement rank framework, including Toeplitz-like ~\cite{sindhwani2015structured}, LDR-SD and LDR-TD matrices~\cite{thomas2018learning}, compare very favorably to other compression approaches. %
Following previous experimental settings~\cite{chen2015compressing,sindhwani2015structured,thomas2018learning}, we compare our proposed classes to several baselines using dense structured matrices to compress the hidden layer of a single hidden layer neural network.
Competing methods include simple low-rank factorizations~\cite{denil2013predicting}, circulant matrices (equivalent to 1-dimensional convolutions)~\cite{cheng2015exploration}, the adaptive Fastfood transform~\cite{yang2015deep}, and low displacement rank methods~\cite{sindhwani2015structured,thomas2018learning} which implicitly define a structured matrix through a displacement equation and admit specialized fast divide-and-conquer algorithms~\cite{desa2018two}.
Our implementation is built on top of the publicly available implementation of~\citet{thomas2018learning} with the same hyperparameters, and we report their numbers for the competing baseline methods directly.
We test on the three main datasets from~\citet{thomas2018learning}: two challenging variants of MNIST---one with randomly rotated images and random background, the other with correlated background noise---and the standard CIFAR-10 dataset. %

\begin{table*}[ht!]
  \centering
  \caption{Test accuracy when replacing the hidden layer with structured classes. For the BPBP methods, the permutations $P$ have been fixed to the bit-reversal permutation.
    The butterfly parameterization achieves higher accuracy than the unstructured
    layer on all datasets.
  }
  \begin{tabular}{@{}lllll@{}}
    \toprule
    \textbf{Method}                                         &  \textbf{MNIST-bg-rot}            & \textbf{MNIST-noise}              & \textbf{CIFAR-10}  & \textit{\textcolor{black}{Compression factor}} \\ \midrule
    Unstructured                                            & 44.08                             & 65.15                             & 46.03   & \textit{\textcolor{black}{1}}                            \\
    \midrule
    BPBP (complex, fixed permutation)                  & \textbf{46.26}                             & 77.00                             & \textbf{49.93} & \textit{\textcolor{black}{39.4}} \\
    BPBP (real, fixed permutation)                     & 46.16                             & 75.00                             & 48.69 & \textit{\textcolor{black}{56.9}}    \\
    \midrule
    LDR-TD ~\cite{thomas2018learning}                       &    45.81                          & \textbf{78.45}                    & 45.33               & \textit{\textcolor{black}{56.9}}          \\
    Toeplitz-like ~\cite{sindhwani2015structured}           & 42.67                             & 75.75                             & 41.78               & \textit{\textcolor{black}{56.9}}                  \\
    Fastfood  ~\cite{yang2015deep}                          & 38.13                             & 63.55                             & 39.64               & \textit{\textcolor{black}{78.7}}                  \\
    Circulant  ~\cite{cheng2015exploration}                 & 34.46                             & 65.35                             & 34.28               & \textit{\textcolor{black}{93.0}}                  \\
    Low-rank ~\cite{denil2013predicting}                    & 35.67                             & 52.25                             & 32.28                   & \textit{\textcolor{black}{56.9}}              \\
    \bottomrule
  \end{tabular}
  \label{table:images}
\end{table*}

Table~\ref{table:images} reports results for variants of our butterfly parametrization,
compared to the unstructured matrix baseline and other structured matrix approaches.
Notably, the butterfly methods achieve higher classification accuracy than
the fully-connected layer on all datasets and are highly competitive with the other approaches.

We note that improvements over unconstrained matrices can arise from lower generalization error due to fewer parameters (relating to VC bounds, Proposition~\ref{prop:vc}), or better inductive bias encoded by the structured class.
For example, convolutions are important in image tasks due to encoding shift equivariance, and \citet{thomas2018learning} hypothesize that their structured classes improve over FC layers through imposing approximate equivariance to more general transformations.
Since our BP parametrization can represent arbitrary convolutions, it can encode these important priors.

\paragraph{ResNet}

In addition to the standard single hidden layer benchmarks, we test the effect of using butterfly layers in a standard ResNet18~\citep{he2016deep} implementation on the CIFAR-10 dataset.
This architecture is normally fully convolutional, ending with a FC layer of dimensions $512 \times 10$ before the softmax.
However, we experiment with adding an additional FC or structured layer right before this final FC layer.
Table~\ref{table:resnet} shows that the ResNet18 architecture can benefit from an additional fully connected layer,
and using a BPBP layer instead improves performance even more while adding a negligible (0.07\% increase) number of parameters to the original model.

\begin{table}[h!]
  \caption{Classification accuracy for the ResNet18 architecture with different layers inserted before the final FC/softmax layer.}
  \centering
\begin{tabular}{@{}llll@{}}
  \toprule
  Last layer & None  & FC    & BPBP \\
  \midrule
  Accuracy   & 93.58 $\pm$ 0.15 & 93.89 $\pm$ 0.19 & \textbf{94.01} $\pm$ 0.09   \\
  \bottomrule
\end{tabular}
\label{table:resnet}
\end{table}

\subsection{Training and Inference Speed Comparison}
\label{sec:speed}

\begin{figure}[ht]
  \centering
  \iftoggle{arxiv}{
    \begin{subfigure}{0.49\linewidth}
      \includegraphics[width=\linewidth]{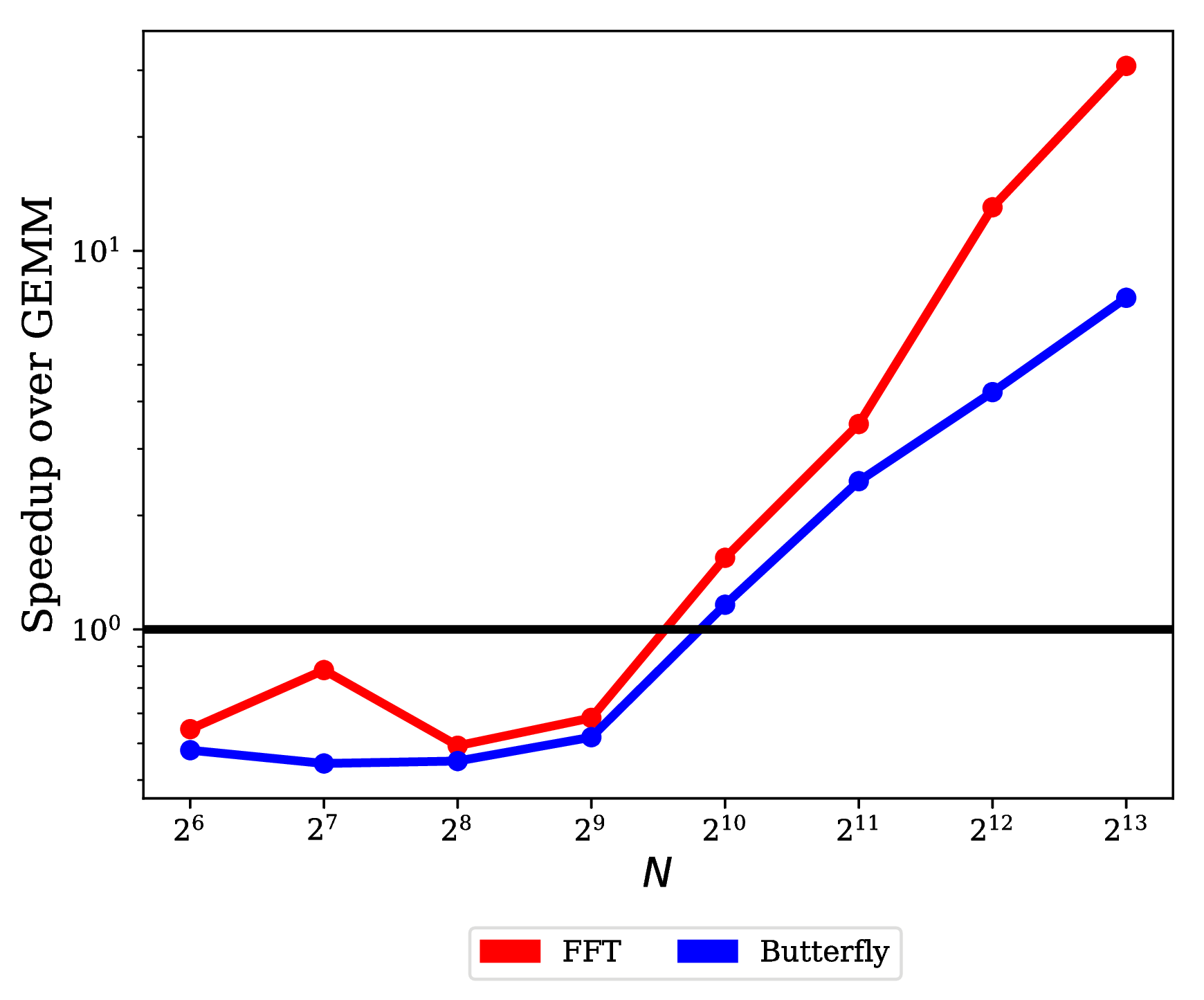}
      \caption{Training}
    \end{subfigure}\hfill%
    \begin{subfigure}{0.49\linewidth}
      \includegraphics[width=\linewidth]{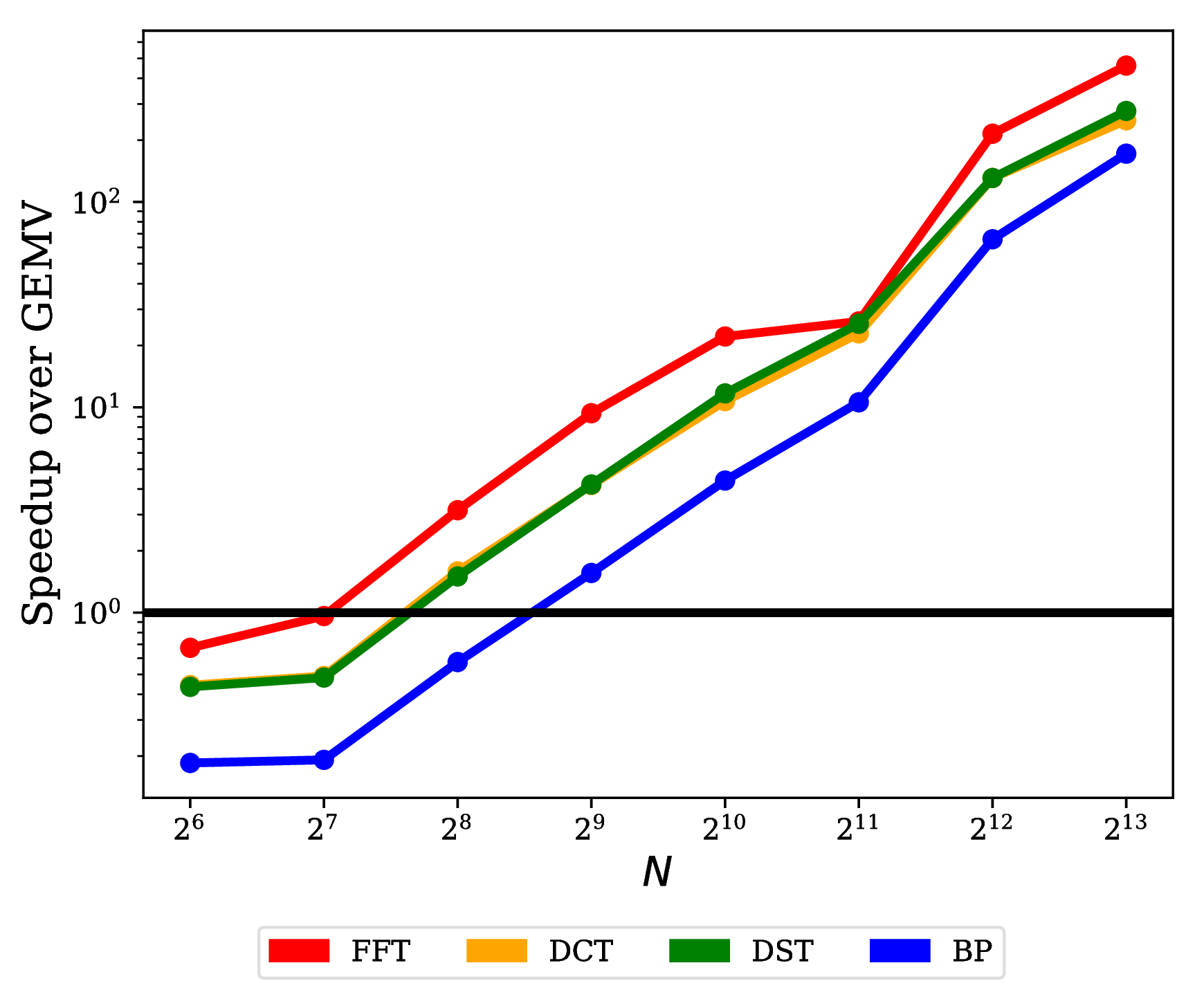}
      \caption{Inference}
    \end{subfigure}
  }
  {
    \begin{subfigure}{0.49\linewidth}
      \includegraphics[width=\linewidth]{figures/speed_training_plot.png}
      \caption{Training}
    \end{subfigure}\hfill%
    \begin{subfigure}{0.49\linewidth}
      \includegraphics[width=\linewidth]{figures/speed_plot.png}
      \caption{Inference}
    \end{subfigure}
  }

  \caption{Speedup of FFT and Butterfly against dense matrix-matrix
    multiply (GEMM) for training, and FFT, DCT, DST, and BP against dense matrix-vector
    multiply (GEMV) for inference. Butterfly's performance is constant with respect to any of the possible transforms it can learn, in contrast to the highly tuned implementations for specific transforms.}
  \label{fig:speed}
\end{figure}

By design, the BP parameterization yields a fast algorithm of complexity
$O(N \log N)$, no matter which transform it learns.
Moreover, given the parameters of the BP model, it is easy to implement this
fast algorithm (this can be done in 5 lines of Python, and our code provides a
function to do this automatically).
The BP parameterization captures many common transforms
(Section~\ref{sec:learning_fast_transforms}), and its implementation makes no
transform-specific optimizations.
Nevertheless, our simple implementation is surprisingly competitive with
hand-tuned kernels both for training and for inference (after the parameters of
the BP model are learned and we wish to evaluate $BPx$ for new input $x$).
In Figure~\ref{fig:speed}, we compare the speed of the BP fast multiplication
against specialized implementation of common transforms such as the FFT, DCT,
and DST (all have complexity $O(N \log N)$), using dense matrix-vector multiply
(GEMV, complexity $O(N^2))$ as a baseline.
For training with realistic input sizes $N = 1024$ and batch size 256 on GPU,
the training time (forward and backward) of butterfly matrix is 15\% faster than
dense matrix multiply (GEMM from cuBLAS) and within 40\% of FFT (from cuFFT).
For inference on CPU, the BP fast multiplication can be one or two orders of
magnitude faster than GEMV, is within a factor of 5 of the FFT, and is within a
factor of 3 of the DCT and the DST, across a range of input sizes.
The GEMM/GEMV and the FFT are two of the most heavily tuned numerical routines.

\section{Conclusion}%
\label{sec:conclusion}

We address the problem of automatically learning fast algorithms for a class of important linear transforms, through a parameterization of recursive algorithms via butterfly factorizations.
We validate our method by learning transforms including the DFT, DCT, Hadamard transform, and convolutions up to machine precision and dimension $N=1024$.
Finally, we show that the same method yields consistent performance improvements and substantial compression and speed increases as a component of end-to-end ML models.

\section*{Acknowledgments}

We thank Maximilian Lam for his help with early experiments.

We gratefully acknowledge the support of DARPA under Nos.\ FA87501720095 (D3M) and FA86501827865 (SDH), NIH under No.\ U54EB020405 (Mobilize), NSF under Nos.\ CCF1763315 (Beyond Sparsity) and CCF1563078 (Volume to Velocity), ONR under No.\ N000141712266 (Unifying Weak Supervision), the Moore Foundation, NXP, Xilinx, LETI-CEA, Intel, Google, NEC, Toshiba, TSMC, ARM, Hitachi, BASF, Accenture, Ericsson, Qualcomm, Analog Devices, the Okawa Foundation, and American Family Insurance, Google Cloud, Swiss Re, and members of the Stanford DAWN project: Intel, Microsoft, Teradata, Facebook, Google, Ant Financial, NEC, SAP, VMWare, and Infosys. The U.S. Government is authorized to reproduce and distribute reprints for Governmental purposes notwithstanding any copyright notation thereon. Any opinions, findings, and conclusions or recommendations expressed in this material are those of the authors and do not necessarily reflect the views, policies, or endorsements, either expressed or implied, of DARPA, NIH, ONR, or the U.S. Government.
Matthew Eichhorn and Atri Rudra’s research is supported by NSF grant CCF-1763481.

\bibliographystyle{icml2019}
\bibliography{refs}

\clearpage

\onecolumn

\appendix

\section{Matrix Factorizations of Linear Transforms}
\label{sec:examples}

Table~\ref{tab:formulas} summarizes the transforms considered in Section~\ref{sec:learning_fast_transforms}.
In general, they transform a (real or complex) vector
$x = [x_0, \dots, x_{N-1}]$ into another (real or complex) vector
$X = [X_0, \dots, X_{N-1}]$ by expressing the input signal in terms of another
set of basis.

\begin{table*}[t]
  \centering
  \caption{Formulas for transforms considered in Section~\ref{sec:learning_fast_transforms} and Figure~\ref{fig:learning_fast_transforms}.}
  \begin{tabular}{@{}ll@{}}
    \toprule
                                                       Transform & Formula \\
    \hline
    DFT       & $X_k = \sum_{n=0}^{N-1} x_n e^{-\frac{i2\pi}{N} nk}$  \\
    DCT       & $X_k = \sum_{n=0}^{N-1} x_n \cos \left[ \frac{\pi}{N} \left( n+ \frac{1}{2} \right) k \right]$  \\
    DST       & $X_k = \sum_{n=0}^{N-1} x_n \sin \left[ \frac{\pi}{N} \left( n+ \frac{1}{2} \right) (k + 1) \right]$  \\
    Convolution & $X_k = \sum_{n=0}^{N-1} x_n g_{k-n}$ \\
    Hadamard & $H_1 = 1$, $H_m = \frac{1}{\sqrt{2}} \begin{bmatrix} H_{m-1} & H_{m-1} \\ H_{m-1} & -H_{m-1} \end{bmatrix}$  \\
    Hartley       & $X_k = \sum_{n=0}^{N-1} x_n \left[ \cos \left( \frac{2\pi}{N} nk \right) + \sin \left( \frac{2\pi}{N} nk \right) \right]$  \\
    \midrule
    Legendre & $X_k = \sum_{n=0}^{N-1} x_n L_k(2n/N-1)$, $L_k(x) = \frac{1}{2^k k!}\frac{d^k}{dx^k}(x^2-1)^k$  \\
    \midrule
    Randn & $(T_N)_{ij} {\sim} \mathcal{N}(1, \frac{1}{N})$  \\
    \bottomrule
  \end{tabular}
  \label{tab:formulas}
\end{table*}

\subsection{Discrete Cosine Transform (DCT) Matrix}
\label{subsec:dct}

The DCT (type II) of a vector $x \in \mathbb{R}^N$ is defined as
\begin{equation*}
  X_k = \sum_{n=0}^{N-1} x_n \cos \left[ \frac{\pi}{N} \left(n + \frac{1}{2}\right) k \right], \qquad k = 0, \dots, N-1.
\end{equation*}

As described in \citet{makhoul1980dct}, the DCT of $x$ can be written in terms of the FFT of order $N$. To do this, we permute $x$ into $v$ by separating the even and odd indices and reversing the odd indices (e.g. \ $[0, 1, 2, 3] \to [0, 2, 3, 1]$), taking the FFT of $v$ to obtain $V$, and multiplying each $V_k$ ($k = 0, \hdots, N-1$) by $2 e^{-\frac{i\pi k}{2N}}$ and taking the real part to get $X_k$.

Written in terms of matrix factorization:
\begin{align*}
  DCT_N = \Re \, \diag{2e^{-\frac{i\pi k}{2N}}} F_N P',
\end{align*}
where $\Re$ takes the real part and $P'$ is a permutation matrix (the permutation
done at the beginning of the DCT).
Recall that $F_N$ has the form 
\begin{equation*}
    F_N = B_{N} \begin{bmatrix} B_{N/2} & 0 \\ 0 & B_{N/2} \end{bmatrix} \dots \begin{bmatrix} B_{2} & \hdots & 0 \\ \vdots & \ddots & \vdots \\ 0 & \hdots & B_2 \end{bmatrix} P,
\end{equation*}
where $P$ is the
bit-reversal permutation matrix. $\diag{2e^{-\frac{i\pi k}{2N}}}$ can be combined with $B_{N}$ to form another
butterfly factor $B_N'$.
Thus the DCT has this factorization:
\begin{equation*}
    DCT_N = \Re \, B_{N}' \begin{bmatrix} B_{N/2} & 0 \\ 0 & B_{N/2} \end{bmatrix} \dots \begin{bmatrix} B_{2} & \hdots & 0 \\ \vdots & \ddots & \vdots \\ 0 & \hdots & B_2 \end{bmatrix} P P'.
\end{equation*}
This is a (BP)${}^2$ factorization (with the additional final step of computing the real part) with the left BP performing the FFT and final scaling, the right butterfly matrix as the identity, and the right permutation matrix as the permutation at the beginning of the DCT.

\subsection{Discrete Sine Transform (DST) Matrix}
\label{subsec:dst}

The DST (type II) of a vector $x \in \mathbb{R}^N$ is defined as
\begin{equation*}
  X_k = \sum_{n=0}^{N-1} x_n \sin \left[ \frac{\pi}{N} \left(n + \frac{1}{2}\right) (k+1) \right], \qquad k = 0, \dots, N-1.
\end{equation*}

Just as with the DCT, we express the DST of $x$ in terms of the FFT of order $N$. First, we permute $x$ into $v$ by separating the even and odd indices and reversing the odd indices (e.g. \ $[0, 1, 2, 3] \to [0, 2, 3, 1]$), then negate those elements in the second half of $v$. We then multiply each $v_k$ with $e^{-\frac{i 2 \pi k}{N}}$. Next, we take the FFT of $v$ to obtain $V$. Finally multiply each $V_k$ ($k = 0, \hdots, N-1$) by $ 2i e^{-\frac{i\pi k}{2N}}$ and take the real part to get $X_k$.

Written in terms of matrix factorization:
\begin{align*}
  DST_N = \Re \, \diag{2i e^{-\frac{i\pi k}{2N}}} F_N D P',
\end{align*}
where $\Re$ takes the real part, $D$ is the matrix $\begin{bmatrix} I_{N/2} & 0 \\ 0 & -I_{N/2} \end{bmatrix} \diag{e^{-\frac{i 2 \pi k}{N}}}$ and $P'$ is a permutation matrix (the permutation
done at the beginning of the DST).
Recall that $F_N$ has the form 
\begin{equation*}
    F_N = B_{N} \begin{bmatrix} B_{N/2} & 0 \\ 0 & B_{N/2} \end{bmatrix} \dots \begin{bmatrix} B_{2} & \hdots & 0 \\ \vdots & \ddots & \vdots \\ 0 & \hdots & B_2 \end{bmatrix} P,
\end{equation*}
where $P$ is the
bit-reversal permutation matrix. We may combine $\diag{2i e^{-\frac{i\pi k}{2N}}}$ with $B_{N}$ to obtain a new butterfly factor, which we call $B_{N}'$.
Thus the DST has this factorization:
\begin{equation*}
    DST_N = \Re \, B_{N}' \begin{bmatrix} B_{N/2} & 0 \\ 0 & B_{N/2} \end{bmatrix} \dots \begin{bmatrix} B_{2} & \hdots & 0 \\ \vdots & \ddots & \vdots \\ 0 & \hdots & B_2 \end{bmatrix} P D P'.
\end{equation*}
Since $D$ is a diagonal matrix
and $P$ is the
bit-reversal permutation matrix, we have that $PD = D'P$ where $D'$ is the diagonal matrix obtained by permuting the diagonals of $D$.
Hence
\begin{equation*}
  DST_N = \Re \, B_{N}' \begin{bmatrix} B_{N/2} & 0 \\ 0 & B_{N/2} \end{bmatrix} \dots \begin{bmatrix} B_{2} & \hdots & 0 \\ \vdots & \ddots & \vdots \\ 0 & \hdots & B_2 \end{bmatrix} D' P P'.
\end{equation*}
The diagonal matrix $D'$ can be combined with the butterfly factor
$\begin{bmatrix} B_{2} & \hdots & 0 \\ \vdots & \ddots & \vdots \\ 0 & \hdots & B_2 \end{bmatrix}$
to yield another butterfly factor of the same form.
Therefore:
\begin{equation*}
  DST_N = \Re \, B_{N}' \begin{bmatrix} B_{N/2} & 0 \\ 0 & B_{N/2} \end{bmatrix} \dots \begin{bmatrix} B'_{2} & \hdots & 0 \\ \vdots & \ddots & \vdots \\ 0 & \hdots & B'_2 \end{bmatrix} P P'.
\end{equation*}
Hence, this factorization of the DST is a (BP)$^2$ factorization (with the
additional final step of computing the real part) with the left BP performing
the FFT and final scaling, the right butterfly matrix as the identity, and the
right permutation matrix as the permutation at the beginning of the DST.

\subsection{Hadamard Matrix}
\label{subsec:hadamard}

The Hadamard matrix (for powers of 2) is defined recursively as $H_1 = 1$, and
$H_{N} = \begin{bmatrix} H_{N/2} & H_{N/2} \\ H_{N/2} & -H_{N/2} \end{bmatrix}$.
Thus we have the recursive factorization:
\begin{equation*}
  H_N = \begin{bmatrix} I_{N/2} & I_{N/2} \\ I_{N/2} & -I_{N/2} \end{bmatrix} \begin{bmatrix} H_{N/2} & 0 \\ 0 & H_{N/2} \end{bmatrix},
\end{equation*}
which is a BP factorization with each butterfly factor, $B_{N/2^k} = \begin{bmatrix} I_{N/2^{k+1}} & I_{N/2^{k+1}} \\ I_{N/2^{k+1}} & -I_{N/2^{k+1}} \end{bmatrix}$ and with permutation matrix $P^{(N)} = I_N$. Here, the entries of the butterfly factors may be real, instead of complex.

\subsection{Convolution}
\label{subsec:convolution}

Here we apply the decomposition of FFT to see if we can learn the decomposition
of fast convolution.

Suppose we have a fixed vector $h \in \mathbb{C}^{N}$ and the linear map we're
interested in is $x \mapsto h * x$.
We can write this convolution with $h$ explicitly as a \emph{circulant} matrix:
\begin{equation*}
  A =
  \begin{bmatrix}
    h_0 & h_{N-1} & \dots & h_2 & h_1 \\
    h_1 & h_0 & h_{N-1} & & h_2 \\
    \vdots & h_1 & h_0 & \ddots & \vdots \\
    h_{N-2} & & \ddots & \ddots & h_{N-1} \\
    h_{N-1} & h_{N-2} & \dots & h_1 & h_0
  \end{bmatrix}.
\end{equation*}
We can compute convolution by the DFT:
\begin{equation*}
  Ax = F_N^{-1} ((F_N h) \odot (F_N x)),
\end{equation*}
where $F_N^{-1}$ denotes the inverse Fourier transform matrix where
$(F_N^{-1}) = \frac{1}{N} \omega_N^{ij}$ and $\odot$ denotes elementwise
multiplication.
Since $h$ is just some fixed vector, elementwise multiplication with
$F_N h$ is just multiplication by some fixed diagonal matrix $D$.
Then
\begin{equation*}
  Ax = F_N^{-1} D F_N x.
\end{equation*}
Note that the inverse Fourier transform has the same algorithm, and thus the same factorization, as the Fourier transform (with different twiddle factors, $\omega_N^{ij}$ instead of
$\omega_N^{-ij}$). Hence, we can express
\begin{equation*}
  A = \frac{1}{N} \tilde{B}_{N} \begin{bmatrix} \tilde{B}_{N/2} & 0 \\ 0 & \tilde{B}_{N/2} \end{bmatrix} \dots \begin{bmatrix} \tilde{B}_{2} & \hdots & 0 \\ \vdots & \ddots & \vdots \\ 0 & \hdots & \tilde{B}_2 \end{bmatrix} P D B_N \begin{bmatrix} B_{N/2} & 0 \\ 0 & B_{N/2} \end{bmatrix} \dots \begin{bmatrix} B_{2} & \hdots & 0 \\ \vdots & \ddots & \vdots \\ 0 & \hdots & B_2 \end{bmatrix} P,
\end{equation*}
where $P$ is the bit-reversal permutation. We may fold the $\frac{1}{N}$ into $\tilde{B}_{N}$ to obtain a new butterfly factor $\tilde{B}_{N}'$, and we may similarly fold the diagonal matrix $D$ into $B_N$ to obtain a new butterfly factor $B_N'$. Hence, our final factorization of convolution / the circulant matrix is :
\begin{equation*}
  A = \tilde{B}_{N}' \begin{bmatrix} \tilde{B}_{N/2} & 0 \\ 0 & \tilde{B}_{N/2} \end{bmatrix} \dots \begin{bmatrix} \tilde{B}_{2} & \hdots & 0 \\ \vdots & \ddots & \vdots \\ 0 & \hdots & \tilde{B}_2 \end{bmatrix} P B_N' \begin{bmatrix} B_{N/2} & 0 \\ 0 & B_{N/2} \end{bmatrix} \dots \begin{bmatrix} B_{2} & \hdots & 0 \\ \vdots & \ddots & \vdots \\ 0 & \hdots & B_2 \end{bmatrix} P,
\end{equation*}
which is a (BP)$^2$ factorization.

Similarly, the skew-circulant matrix also lies in (BP)$^2$:
\begin{equation*}
  A =
  \begin{bmatrix}
    h_0 & -h_{N-1} & \dots & -h_2 & -h_1 \\
    h_1 & h_0 & -h_{N-1} & & -h_2 \\
    \vdots & h_1 & h_0 & \ddots & \vdots \\
    h_{N-2} & & \ddots & \ddots & -h_{N-1} \\
    h_{N-1} & h_{N-2} & \dots & h_1 & h_0
  \end{bmatrix}.
\end{equation*}

\subsection{Toeplitz Matrices}
\label{subsec:toeplitz}

Let $T_N$ be the Toeplitz matrix:
\begin{equation*}
    T_N = \begin{bmatrix} 
        t_0 & t_{-1} & \hdots & t_{-N+2} & t_{-N+1} \\
        t_1 & t_0 & t_{-1} &  & t_{-N+2} \\
        \hdots & t_1 & t_0 & \ddots & \hdots \\
        t_{N-2} &  & \ddots & \ddots & t_{-1} \\
        t_{N-1} & t_{N-2} & \hdots & t_1 & t_0   \end{bmatrix}.
\end{equation*}

Define $\tilde{T}_N$ to be:
\begin{equation*}
    \tilde{T}_N = \begin{bmatrix} 
        0 & t_{N-1} & \hdots & t_{2} & t_{1} \\
        t_{-N+1} & 0 & t_{N-1} &  & t_{2} \\
        \hdots & t_{-N+1} & 0 & \ddots & \hdots \\
        t_{-2} &  & \ddots & \ddots & t_{N-1} \\
        t_{-1} & t_{-2} & \hdots & t_{-N+1} & 0   \end{bmatrix}.
\end{equation*}

Then, $T_N = \begin{bmatrix} I_N & 0 \end{bmatrix} \begin{bmatrix} T_N & \tilde{T}_N \\ \tilde{T}_N & T_N \end{bmatrix} \begin{bmatrix} I_N \\ 0 \end{bmatrix}$. Note that the inner matrix is a $2N \times 2N$ circulant matrix that can be decomposed into a (BP)${}^2$ factorization as described in \ref{subsec:convolution}. Therefore, our final factorization for Toeplitz matrices is contained within (BP)$^2_2$. 

\subsection{Orthogonal Polynomial Matrices}
\label{subsec:ops}

Although the ability to represent general orthogonal polynomial matrices in terms of butterfly matrices is left as an open problem, we nonetheless present an alternate sparse factorization.

\begin{definition} \label{def:op}
    A family of polynomials $\{p\} = p_0(x), p_1(x), \hdots \in \mathbb{R}[x]$ is \textit{orthogonal} over $\mathbb{R}$ if:
    \begin{itemize}
      \item $p_0(x) = c_1$
      \item $p_1(x) = a_1x + b_1$
      \item $p_i(x) = (a_ix + b_i)p_{i-1}(x) + c_i \, p_{i-2}(x)$ for all $i \geq 2$
    \end{itemize}
    We say that $\{p\}$ is parameterized by real sequences $\{ a_i, b_i, c_i : i \in \mathbb{N} \}$ (with $c_1$ and each $a_i \in \mathbb{R}\setminus \{0\}$).
\end{definition}

\begin{definition} \label{def:opm}
    Given a family of orthogonal polynomials $\{p\}$, we may define the \textit{orthogonal polynomial matrix} $P_{[s:n]} \in \mathbb{R}^{(n-s) \times n}$ such that:
    \begin{equation*}
        p_{s+i} = \sum_{j=0}^{n} \left(P_{[s:n]}\right)_{ij} x^j,
        \hspace{20pt}
        0 \leq i < n-s
    \end{equation*}
\end{definition}

For sake of clarify, we formulate the decomposition using matrices of polynomials. We note that each polynomial entry with degree bounded by $d$ may be expanded into a $d \times 2d$ Toeplitz convolution matrix if one desires matrices of real coefficients.

For a given family of orthogonal polynomials $\{p\}$ parameterized by $\{a_j,b_j,c_j : 1 \leq j \leq n-1\}$, let $T_j \in \mathbb{R}[x]^{2 \times 2}$ be a \textit{transition matrix} defined by:
\[
    \begin{bmatrix} a_j x + b_j & c_j \\ 1 & 0 \end{bmatrix}.
\]
For convenience of notation, let $T_0 = I$. Let $T_{[\ell,r]} \in \mathbb{R}[x]^{2 \times 2}$ be a \textit{transition product matrix} defined by:
\[
    T_{[\ell:r]} \equiv T_{\ell} \cdot T_{(\ell-1)} \hdots T_{(r+1)} \cdot T_{r} \equiv
    \begin{bmatrix}
        A_{[\ell:r]}(x) &  B_{[\ell:r]}(x) \\  C_{[\ell:r]}(x) &  D_{[\ell:r]}(x)
    \end{bmatrix}.
\]
From these definitions, we see that for all $j \geq 0$,
\[
    \begin{bmatrix} p_{j+1}(x) \\ p_j(x) \end{bmatrix} 
    = T_j \begin{bmatrix} p_j(x) \\ p_{j-1}(x) \end{bmatrix}
    = T_{[j:0]} \begin{bmatrix} p_1(x) \\ p_0(x) \end{bmatrix}.
\]

We use this to formulate the following decomposition of the orthogonal polynomial matrix $P_{[0:n]}$.
\begin{align}
    P_{[0:n]} = 
    \dimmatrix{
         p_0(x) \\ p_1(x) \\ \vdots \\ p_{n-1}(x)
    }{n \times 1} = 
    \dimmatrix{ 
        0 & 1 & 0 & 0 & \hdots & 0 & 0 \\
        0 & 0 & 0 & 1 & \hdots & 0 & 0 \\ 
        \vdots & \vdots &  \vdots &  \vdots & \ddots &  \vdots &  \vdots \\
        0 & 0 & 0 & 0 & \hdots & 0 & 1 
    }{n \times 2n}
    \dimmatrix{
        T_{[0:0]} \\ T_{[1:0]} \\ \vdots \\ T_{[n-1:0]}
    }{2n \times 2}
    \dimmatrix{
        p_1(x) \\ p_0(x)
    }{2 \times 1}.
\end{align}

The first ``stretched" identity matrix serves the function of selecting every other entry from the vector of $2n$ polynomials to its right. We focus our attention on the middle matrix. Noting that $T_{[\ell:r]}$ = $T_{[\ell:m]} \cdot T_{[m-1:r]}$ for any $r \leq m \leq \ell$, we may represent this block matrix as:
\begin{align}
    \dimmatrix{
        T_{[0:0]} \\ T_{[1:0]} \\ \vdots \\ T_{[n-1:0]}
    }{2n \times 2}
    =
    \dimmatrix{
        T_{[0:0]} \\ \vdots \\ T_{[\frac{n}{2}-1:0]} \\ \bigzero{n \times 2} 
    }{2n \times 2}
    + 
    \dimmatrix {
        \bigzero{n \times 2} \\ T_{[\frac{n}{2}:\frac{n}{2}]} \\ \vdots \\ T_{[n-1:\frac{n}{2}]}
    }{2n \times 2}
    \dimmatrix{
        T_{[\frac{n}{2}-1:0]}
    }{2 \times 2}
    = 
    \dimmatrix{
        \begin{matrix} T_{[0:0]} \\ \vdots \\ T_{[\frac{n}{2}-1:0]} \end{matrix} & 
        \bigzero{n \times 2} \\
        \bigzero{n \times 2} & 
        \begin{matrix} T_{[\frac{n}{2}:\frac{n}{2}]} \\ \vdots \\ T_{[n-1:\frac{n}{2}]} \end{matrix}
    }{2n \times 4}
     \dimmatrix{
        \bigeye{2 \times 2} \\
        T_{[\frac{n}{2}-1:0]}
    }{4 \times 2}.
\end{align}

Notice that the left matrix in this last expression consists of two matrices with the same structure as the first expression, but of half the size. Hence, we may repeat the same decomposition on each of the sub-matrices. In general, the decomposition becomes:

\begin{align}
    \dimmatrix{ 
        T_{[0:0]} \\ T_{[1:0]} \\ \vdots \\ T_{[n-1:0]} 
    }{2n \times 2}
    = 
    \dimmatrix{
        \bigeye{2 \times 2} & \bigzero{2 \times 2} & \hdots & \bigzero{2 \times 2} \\
        T_1 & \bigzero{2 \times 2} & \hdots & \bigzero{2 \times 2} \\
        \bigzero{2 \times 2} & \bigeye{2 \times 2} & \hdots & \bigzero{2 \times 2} \\
        \bigzero{2 \times 2} & T_3 & \hdots & \bigzero{2 \times 2} \\
        \vdots & \vdots & \ddots & \vdots \\
        \bigzero{2 \times 2} & \bigzero{2 \times 2} & \hdots & \bigeye{2 \times 2} \\
        \bigzero{2 \times 2} & \bigzero{2 \times 2} & \hdots & T_{n-1}
     }{2n \times n}
     \hdots 
     \dimmatrix{
        \bigeye{2 \times 2} & \bigzero{2 \times 2} \\
        T_{[\frac{n}{4}-1:0]} & \bigzero{2 \times 2} \\
        \bigzero{2 \times 2} & \bigeye{2 \times 2} \\
        \bigzero{2 \times 2} & T_{[\frac{3n}{4}-1:\frac{n}{2}]}
    }{8 \times 4}
    \dimmatrix{
        \bigeye{2 \times 2} \\
        T_{[\frac{n}{2}-1:0]}
    }{4 \times 2}.
\end{align}

\paragraph{Discrete Legendre Transform}

The Discrete Legendre Transform (DLT) of a vector $x \in \mathbb{R}^N$ is defined as:

\begin{equation*}
    X_k = \sum_{n=0}^{N-1} x_n L_k\left(\frac{2n}{N-1}\right),
\end{equation*}

where $L_k$ is the $k$'th Legendre polynomial. The Legendre polynomials are a family of orthogonal polynomials with:

\begin{equation*}
    L_0(x) = 1 \hspace{30pt} L_1(x) = x \hspace{30pt} L_k(x) = \left(\tfrac{2k-1}{k} \right)x L_{k-1}(x) - \left(\tfrac{k-1}{k}\right) L_{k-2}(x), \hspace{10pt} k \geq 2.
\end{equation*}

Hence, the DLT may be factored as described above. 

\section{Proofs}%
\label{sec:proofs}

\subsection{VC Dimension Bound for Neural Network with Butterfly Layers}
\label{subsec:vc_dim}

\begin{proposition}\label{prop:vc}
Let $\mathcal{F}$ denote the class of neural networks with $L$ layers, each is a
butterfly layer using the BP or BPBP parameterization, with fixed permutation,
$W$ total parameters, and piecewise linear activations.
Let $\sign \mathcal{F}$ denote the corresponding classification functions, i.e.
$\{x \mapsto \sign f(x) : f \in \mathcal{F}\}$.
The VC dimension of this class is
\begin{equation*}
  \VCdim(\sign \mathcal{F}) = O(L W \log W).
\end{equation*}
\end{proposition}

Because the parameters within a layer interact multiplicatively, the standard VC
dimension bound for fully-connected
layers~\citep{bartlett1999almost,bartlett2017nearly} does not apply directly.
However, a variant of the same argument applies to the case where degree of
multiplicative interaction is not too high~\citep[Theorem 3]{thomas2018learning}.

We provide a short proof of the VC dimension bound for neural networks with BP
or BP$^2$ layers based on this result.
\begin{proof}
  Theorem 3 of \citet{thomas2018learning} requires that the entries of the
  linear layer, as polynomials of the parameters, has degree at most
  $c_1 m_l^{c_2}$ for some universal constant $c_1, c_2 > 0$, where $m_l$ is the
  size of output of the $l$-th layer.
  In our case, the BP or BPBP parameterization with fixed permutation has total
  degree at most $2\log_2 n$ in the parameters of $B$, where $n$ is the size of
  the layer, since each $B^{(n)}$ is a product of $\log_2 n$ matrices.
  It thus satisfies the condition of the theorem, and so the VC dimension is
  bounded to be almost linear in the number of parameters:
  \begin{equation*}
    \VCdim(\sign \mathcal{F}) = O(L W \log W).
  \end{equation*}
\end{proof}

\subsection{Proposition ~\ref{prop:expressivity}}

\begin{proof}
\begin{enumerate}
    \item The inclusion of the DFT in (BP)$^1$ is shown in the Case study in Section~\ref{subsec:dft}. The inverse Fourier Transform has the same structure except the twiddle factors of the form $\omega_N^{-ij}$ are replaced with $\omega_N^{ij}$ and all entries of the first butterfly factor are scaled by $\frac{1}{N}$.
    \item The inclusion of the Hadamard Transform in (BP)$^1$ is shown in Section~\ref{subsec:hadamard}.
    \item The inclusion of the DCT in (BP)$^2$ is shown in Section~\ref{subsec:dct}.
    \item The inclusion of the DST in (BP)$^2$ is shown in Section~\ref{subsec:dst}.
    \item The inclusion of the convolution in (BP)$^2$ is shown in Section~\ref{subsec:convolution}.
    \item The inclusion of all $N \times N$ matrices in (BP)$_2^{4N+10}$ follows from the fact that every $N \times N$ matrix may be expressed by a product of at most $2N + 5$ Toeplitz matrices \cite{ye2016toep}. From Section \ref{subsec:toeplitz}, we may conclude that all Toeplitz matrices are in (BP)$^2_2$. Therefore, by appending the BP modules from each Toeplitz matrix, we see that a total of $4N + 10$ BP modules are needed. By left multiplying each butterfly factor by the $2N \times 2N$ diagonal matrix with 1s in the upper half and 0s in the lower half, we ensure that the upper left $N \times N$ entries of the final product are exactly the product of the upper left $N \times N$ entries of each BP module, as required. This diagonal matrix may be absorbed into the adjacent butterfly factor. Hence, the factorization is in (BP)$_2^{4N+10}$.
\end{enumerate}

\end{proof}

\section{Experimental Details and Results}
\label{sec:extraexps}

\subsection{Recovering Fast Transforms}
\label{subsec:learning_fast_transforms_details}

In Section~\ref{sec:learning_fast_transforms}, given a matrix representation of
a transform, we use the BP or BPBP parameter to recover a fast algorithm to the
transform.
We report in Table~\ref{tab:learning_fast_transforms_numbers} the root mean
square error (RMSE) $\sqrt{\frac{1}{N^2} \norm{T_N - B^{(N)}P^{(N)}}}$ for
different transforms and for different values of $N$.

\begin{table*}[t]
  \centering
  \caption{RMSE of learning fast algorithms for common transforms, where we stop early when RMSE $<$ 1e-4.}
  \begin{tabular}{@{}lllllllll@{}}
    \toprule
                                                       Transform                               & N     = 8   & 16  & 32 & 64 & 128 & 256 & 512 & 1024 \\
    \hline
    DFT       & 3.1e-06 & 4.6e-06 & 8.7e-06 & 1.0e-05 & 2.0e-05 & 3.8e-05 & 8.0e-05 & 5.7e-05 \\
    DCT       & 4.4e-06 & 1.1e-05 & 8.6e-06 & 1.2e-05 & 2.1e-05 & 1.9e-05 & 3.1e-05 & 7.3e-05 \\
    DST       & 1.1e-06 & 7.5e-06 & 4.6e-05 & 5.1e-05 & 3.0e-05 & 2.1e-05 & 3.6e-05 & 4.6e-05 \\
    Convolution & 4.0e-06 & 2.5e-05 & 6.4e-05 & 7.6e-05 & 5.9e-05 & 7.8e-05 & 6.3e-05 & 1.9e-02 \\
    Hadamard & 8.8e-07 & 7.8e-06 & 1.3e-05 & 3.9e-05 & 3.5e-05 & 4.5e-05 & 6.1e-05 & 3.6e-05 \\
    Hartley   & 3.4e-06 & 9.0e-06 & 1.1e-05 & 1.3e-05 & 3.6e-05 & 4.3e-05 & 4.5e-05 & 3.6e-05 \\
    \hline
    Legendre & 3.4e-02 & 2.9e-02 & 2.4e-02 & 1.4e-02 & 7.9e-03 & 4.5e-03 & 2.6e-03 & 1.6e-03 \\
    \hline
    Randn & 1.4e-01 & 1.6e-01 & 1.4e-01 & 1.1e-01 & 8.4e-02 & 6.1e-02 & 4.4e-02 & 3.1e-02 \\
    \bottomrule
  \end{tabular}
  \label{tab:learning_fast_transforms_numbers}
\end{table*}

We use Hyperband~\citep{li2017hyperband} to tune the hyperparameters, which
include the learning rate (from 0.0001 to 0.5), initialization, and whether to
share the logits in the permutation block $P^{(N)}$.

\subsection{Fully connected network}

The model is a network with a single hidden layer of dimensions $N \times N$, where $N$ is the input dimension, followed by a fully-connected softmax layer.
We build on top of the framework of~\citet{thomas2018learning}\footnote{Available at \url{https://github.com/HazyResearch/structured-nets}},
replacing the unconstrained or structured matrix with our PyTorch BPBP implementation.
The CIFAR-10 dataset is a grayscale version of input size 1024 since the single hidden layer architecture receives a single channel as input.
With the exception of learning rate, hyperparameters such as batch size 50, validation set comprising 15\% of training data, and fixed momentum at 0.9 are
fixed as reported in Appendix F.1 of their paper.
For the BP methods, the learning rate was tested for the values $\{0.005, 0.01, 0.02, 0.05, 0.1, 0.2\}$; parameters outside this range were found to be ineffective.
For each method, Table~\ref{table:images} reports the test accuracy of the model with the highest validation accuracy.

\subsection{Resnet}

We build on top of the standard ResNet18 model from PyTorch.%
\footnote{Available at \url{https://github.com/pytorch/vision/blob/master/torchvision/models/resnet.py}}
The model is modified for CIFAR-10 by reducing the kernel size and stride for the initial convolution to $3$ and $1$ respectively,
and removing the first max pool layer.
Weight decay of $\lambda=0.0002$ was used.
The learning rate was initialized in $\{0.1, 0.2\}$, and decayed by $\{0.1,0.2\}$ every 25 epochs for 100 epochs total.
For each method, Table~\ref{table:resnet} reports the mean and standard deviation of the test accuracies for the hyperparameters with the highest average validation accuracy.

\subsection{Speed Comparison}
\label{subsec:speed_details}

In Section~\ref{sec:speed}, we benchmark the speed of training and inference of
butterfly factorizations.

For training, we compare our CUDA implementation of the fast algorithm for
butterfly matrices with dense matrix-matrix multiply (GEMM from cuBLAS) and FFT
(from cuFFT).
The batch size is 256, and we measure the total time of the forward and backward
pass.
The experiment is run on a Tesla P100 GPU with 16GB of memory.

For inference, we compare our simple Python implementation of the fast algorithm
for the BP parameterization, against dense matrix-vector multiplication (GEMV),
FFT, DCT, and DST.
Our BP parameterization here refers to the product of a butterfly matrix
$B^{(N)}$ and a fixed permutation $P^{(N)}$ (say, learned from data).
We use the standard dense matrix-vector multiplication implementation in Numpy
(BLAS binding), the FFT implementation from Numpy and the DCT
and DST implementation from Scipy (FFTPACK binding).
We compare their speed in single-threaded mode, running on a server Intel Xeon
CPU E5-2690 v4 at 2.60GHz.

Results are shown in Figure~\ref{fig:speed}.

\section{BP Hierarchy}
\label{sec:bp-conj}

In Definition~\ref{def:bp-hierarchy}, we defined the notion of a BP hierarchy, which we believes captures a natural class of matrices.
To this point, we offer the following observations, the latter left as a conjecture, about the expressiveness of this hierarchy, supplementing the inclusion results of Proposition~\ref{prop:expressivity}.

\begin{proposition}
    For every fixed $c \geq 1$, there is a sufficiently large $N$ such that there is an $N \times N$ matrix $M_N$ that is in (BP)$^{c+1}$ but not in (BP)$^{c}$. 
\end{proposition}

\begin{proof}
    Given $c$, fix $N$ such that $N$ is even and such that $c < \frac{N}{8 \log_2 N}$. For sake of contradiction, assume that every $N \times N$ matrix in (BP)$^{c+1}$ is also in (BP)$^{c}$. Let $A$ be an arbitrary $\frac{N}{2} \times \frac{N}{2}$ matrix. Then, from Proposition~\ref{prop:expressivity}, $A$ is in (BP)$^{2N+10}_{2}$. Therefore, from Definition~\ref{def:bp-hierarchy}, there is some $N \times N$ matrix $M \in$ (BP)$^{2N+10}$ such that the upper-left $\frac{N}{2} \times \frac{N}{2}$ entries are $A$. From our assumption, we can replace the first $c+1$ BP factors in $M$ with $c$ (possibly different) BP factors. We can repeat this process until we are left with $c$ (BP) factors, so $M$ in (BP)$^c$. This representation for $M$ has $c \cdot 2 N \log_2 N$ parameters, which must be less than $\frac{N}{8 \log_2 N} \cdot 2 N \log_2 N = \frac{N^2}{4}$ based on how we fixed $N$ above. However, $A$ (and therefore $M$) has $\frac{N^2}{4}$ arbitrary entries, contradicting that it can be represented with fewer than $\frac{N^2}{4}$ parameters. Hence, there must be some $N \times N$ matrix in (BP)$^{c+1}$ that is not in (BP)$^{c}$.
\end{proof}

\begin{conjecture}
    Let $M$ be an $N \times N$ matrix such that for any $x \in \mathcal{F}^N$, $Mx$ can be computed with an arithmetic circuit of size $N \poly\log(N)$ and depth $\poly\log(N)$. Then, $M$ is in (BP)$_{O(1)}^{\poly\log{N}}$. 
\end{conjecture}
We believe that we can prove an approximation of the above using known approximations of the Jacobi transform by the DCT (up to some scaling)~\cite{szego}. It is known that such transforms have an arithmetic circuit of the kind mentioned in the conjecture above~\cite{driscoll}.
\end{document}